\pdfoutput=1

\documentclass[11pt]{article}

\usepackage[final]{acl}

\usepackage{times}
\usepackage{latexsym}

\usepackage[T1]{fontenc}

\usepackage[utf8]{inputenc}

\usepackage{microtype}

\usepackage{inconsolata}

\usepackage{graphicx}
\usepackage{times}
\usepackage{soul}
\usepackage{url}
\usepackage[utf8]{inputenc}
\usepackage{algorithmic}
\usepackage[switch]{lineno}
\usepackage{hyperref}
\usepackage{url}
\usepackage{amsmath}
\usepackage{amssymb}
\usepackage{graphicx}
\usepackage{booktabs}
\usepackage{multirow}
\usepackage{wrapfig}
\usepackage{makecell}
\usepackage[normalem]{ulem}
\useunder{\uline}{\ul}{}
\usepackage[dvipsnames]{xcolor}         

\title{\textit{Is Cognition Consistent with Perception?}\\ Assessing and Mitigating Multimodal Knowledge Conflicts
in\\Document Understanding}

\author{Zirui Shao$^{1}$\thanks{\,\, Equal contribution.}, 
Feiyu Gao$^{2}$\footnotemark[1],
Zhaoqing Zhu$^{2}$\footnotemark[1], 
Chuwei Luo$^{2}$\thanks{\,\, Corresponding author.}, \\ 
\textbf{Hangdi Xing}$^{1}$,
\textbf{Zhi Yu}$^{1, 3}$\footnotemark[2], 
\textbf{Qi Zheng}$^{2}$,
\textbf{Ming Yan}$^{2}$,
\textbf{Jiajun Bu}$^{1}$ \\
$^{1}$ Zhejiang Key Laboratory of Accessible Perception and Intelligent Systems, \\ Zhejiang University \\ $^{2}$Alibaba Group, \\ $^{3}$ Hangzhou High-Tech Zone (Binjiang) Institute of Blockchain and DataSecurity \\
\texttt{\{shaozirui, yuzhirenzhe\}@zju.edu.cn, feiyu.gfy@alibaba-inc.com}  \\\texttt{\{zzhaoqing.z, luochuwei\}@gmail.com}}

\begin{document}
\maketitle
\begin{abstract}
Multimodal large language models (MLLMs) have shown impressive capabilities in document understanding, a rapidly growing research area with significant industrial demand.
As a multimodal task, document understanding requires models to possess both perceptual and cognitive abilities.
However, due to different types of annotation noise in training, current MLLMs often face conflicts between perception and cognition.
Taking a document VQA task (cognition) as an example, an MLLM might generate answers that do not match the corresponding visual content identified by its OCR (perception).
This conflict suggests that the MLLM might struggle to establish an intrinsic connection between the information it ``sees'' and what it ``understands''.
Such conflicts challenge the intuitive notion that cognition is consistent with perception, hindering the performance and explainability of MLLMs.
In this paper, we define the conflicts between cognition and perception as \textit{Cognition and Perception (C\&P) knowledge conflicts}, a form of multimodal knowledge conflicts, and systematically assess them with a focus on document understanding.
Our analysis reveals that even GPT-4o, a leading MLLM, achieves only 75.26\% C\&P consistency.
To mitigate the C\&P knowledge conflicts, we propose a novel method called \textit{Multimodal Knowledge Consistency Fine-tuning}. 
Our method reduces C\&P knowledge conflicts across all tested MLLMs and enhances their performance in both cognitive and perceptual tasks.
\end{abstract}

\begin{figure*}[t]
    \centering
    \includegraphics[width=0.90\linewidth]{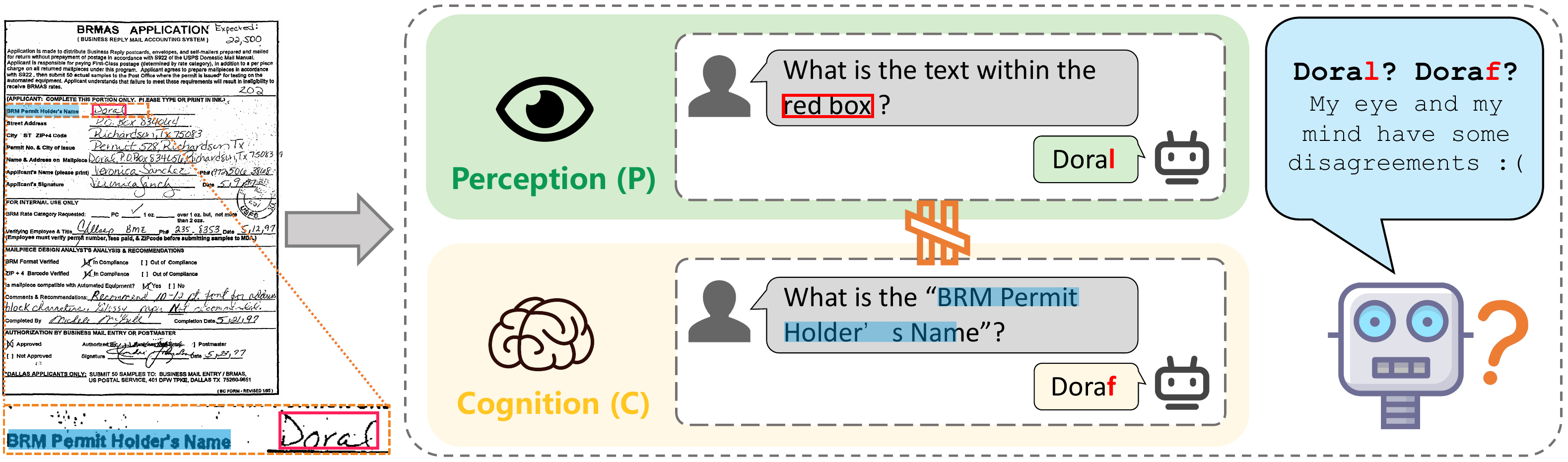} 
    \caption{GPT-4o generates a VQA (cognition) answer that conflicts with the corresponding visual content identified by its OCR (perception). We refer to these multimodal knowledge conflicts in MLLMs as \textit{Cognition and Perception (C\&P) knowledge conflicts}.}
    \label{fig:toyexample}
\end{figure*}

\section{Introduction}
In recent years, multimodal large language models (MLLMs) \citep{gpt4v,team2023gemini,gpt4o,internvl,bai2025qwen2,ye2024mplug,li2024llava} have witnessed rapid development and have demonstrated remarkable capabilities across a wide range of multimodal tasks \citep{antol2015vqa,docvqa,hossain2019comprehensive}.
Of particular note is their application in document understanding \citep{cui2021document,xu2020layoutlm,xu2021layoutlmv2,huang2022layoutlmv3,luo2023geolayoutlm}, an area of high academic and industrial value, where significant progress has been made \citep{zhang2023llavar,ye2023mplugdocowl,ye2023ureader,layoutllm,wang2023docllm,docowl15}.

As a multimodal task, document understanding requires models to accurately perceive visual content (perception, e.g., OCR) and then generate coherent responses (cognition, e.g., VQA) based on that perception.  However, current MLLMs train perception and cognition using different sources of annotation \citep{qwen-vl,docowl15}. Perception typically relies on external OCR engines, while cognition often depends on human-annotated or LLM-generated data \citep{docvqa,dude}. This discrepancy leads to different noise profiles, creating conflicts between perception and cognition. As shown in Figure \ref{fig:toyexample}, GPT-4o \citep{gpt4o} recognizes the text in a certain region of an image as “Doral” but responds to a related VQA question with the text “Doraf”. 
This conflict suggests that GPT-4o struggles to establish a consistent connection between what it “sees” and what it “understands”.
Statistical analysis further underscores this issue, as Figure~\ref{fig:pattern} demonstrates that leading MLLMs like GPT-4o achieve only 75.26\% consistency between perception and cognition (Section \ref{sec:models}).

In this paper, we define intrinsic conflicts between cognitive knowledge and perceptual knowledge within MLLMs, which result in inconsistencies in responses related to cognition and perception, as \textit{Cognition and Perception (C\&P) knowledge conflicts} (Section \ref{sec:definition}). These conflicts undermine the explainability of MLLMs, as they challenge the intuitive notion that cognition is consistent with perception.  Unlike previous research on multimodal knowledge conflicts (e.g., hallucination) \citep{zhai2024halleswitch,li2023evaluating,guan2024hallusionbench,liu2023mitigating}, which focuses solely on conflicts within either cognition or perception, we highlight, for the first time, the conflicts that arise between the two.

We systematically evaluate C\&P knowledge conflicts in the five current MLLMs (Section \ref{sec:models}), focusing on document understanding. For documents, the primary perception task is the recognition of optical characters, while the primary cognitive task is the comprehension of text content. Therefore, we select OCR as the perceptual task and document-related VQA as the cognitive task. To ensure the validity of our evaluation, we eliminate potential confounding factors, such as model failures in following instructions. The experimental results reveal substantial C\&P knowledge conflict in current MLLMs, highlighting the need to resolve these conflicts.
To address this, we introduce a novel method called \textit{Multimodal Knowledge Consistency Fine-tuning}. This method aims to strengthen the connection between cognitive and perceptual tasks through two key components (Section \ref{sec:method}).
First, a special token called \textit{C\&P Link Token} is introduced as a prompt prefix and suffix to connect cognitive and perceptual knowledge.
Second, we design a \textit{C\&P Connector} that guides the model to cross-verify cognitive knowledge using perceptual knowledge.

Comprehensive experiments are conducted on three open-source MLLMs across two series and two parameter sizes. The results indicate that multimodal knowledge consistency fine-tuning improves C\&P consistency (Section \ref{sec:main_results}). Notably, our method also enhances MLLM performance in both cognitive and perceptual tasks (Section \ref{sec:performance_c_p}). This suggests that reducing C\&P knowledge conflicts allows the model to better integrate perceptual and cognitive knowledge, thereby improving its overall capabilities.

Our main contributions are as follows:
\begin{itemize}

    \item To the best of our knowledge, we are the first to identify and introduce the concept of \textit{Cognition and Perception knowledge conflicts}, a form of multimodal knowledge conflicts, in MLLMs.

    \item A systematic evaluation is conducted on current MLLMs to assess the Cognition and Perception knowledge conflicts in document understanding, showing that such conflicts are commonly present in current MLLMs.
    
    \item A novel method called \textit{Multimodal Knowledge Consistency Fine-tuning} is introduced to mitigate the C\&P knowledge conflicts in current MLLMs. Extensive experiments on five public document understanding benchmarks in three MLLMs demonstrate the effectiveness of the proposed method.

\end{itemize}

\section{Problem Statement}

\subsection{The Definition of Cognition and Perception Knowledge Conflicts}
\label{sec:definition}

For a given MLLM \( f(\cdot) \), an image \( x_I \), and a pair of queries consisting of a cognitive query \( x_C \) and a perceptual query \( x_P \), we denote the ground truth for this pair as \( GT \). 
The MLLM's responses for cognitive and perceptual tasks are represented as \( y_C = f(x_C, x_I) \) and \( y_P = f(x_P, x_I) \), respectively.

In the training process of current MLLMs, annotations for perceptual tasks (e.g., OCR) and cognitive tasks (e.g., VQA) are often derived from different sources. 
For example, in the widely used DocVQA dataset \citep{docvqa}, OCR annotations are generated by commercial OCR solutions, while VQA annotations are crowd-sourced.
Differences in annotation origins introduce discrepancies in noise and content, resulting in inconsistent bias that creates conflicts between cognitive and perceptual knowledge, referred to as \textit{Cognition and Perception (C\&P) knowledge conflicts}.
Such conflicts manifest when \( y_C \) and \( y_P \) are inconsistent, i.e., \( \delta(y_C, y_P) = 0 \).
It is important to note that C\&P knowledge conflicts do not consider whether \( y_C = GT \) or \( y_P = GT \). 
To quantify the severity of these conflicts, we introduce \textit{C\&P consistency}. Let \( N \) denote the number of query pairs, with the C\&P consistency calculated as follows:

\begin{equation}
    \text{C\&P Consistency} = \frac{\sum_{i=1}^{N} \delta(y_{C_i}, y_{P_i})}{N}.
\end{equation}

In this paper, we focus on document understanding and follow common practice \citep{fu2024mmecomprehensiveevaluationbenchmark,chen2024pca} by using OCR as a representative perceptual task and VQA as a representative cognitive task. Specifically, given a text \( GT \) within \( x_I \) bounded by \( Box \), \( x_C \) is a VQA query using \( GT \) as the answer, and \( x_P \) is an OCR query operating solely within \( Box \).  In practice, \( Box \) may contain additional text besides \( GT \). Consequently, C\&P knowledge conflicts occur when \( y_P \) does not fully contain \( y_C \). The \( \delta(y_C, y_P) \) can be specifically defined as follows:
\begin{equation}
    \delta(y_C, y_P) = 
    \begin{cases} 
    1, & \text{if } y_C \subseteq y_P \\
    0, & \text{if } y_C \nsubseteq y_P 
    \end{cases}.
\end{equation}

Furthermore, performance gaps may cause models to exhibit C\&P inconsistency. For example, MLLMs may fail to comprehend VQA questions. Therefore, we introduce an auxiliary metric, called ``\textit{Idealized C\&P Consistency},'' which evaluates inconsistencies only when both $ANLS(y_C, GT)$ and $ANLS(y_P, GT)$ are at least 0.5. The ANLS metric~\citep{biten2019scene} is widely used in document understanding to measure text similarity on a scale from 0 to 1. Generally, cases with ANLS below 0.5 are considered complete failures of the model's response to a query. By filtering out these poor cases caused by model performance, this metric provides additional insight into the C\&P consistency under ideal conditions.

\begin{figure*}[t]
    \centering
    \includegraphics[width=1\textwidth]{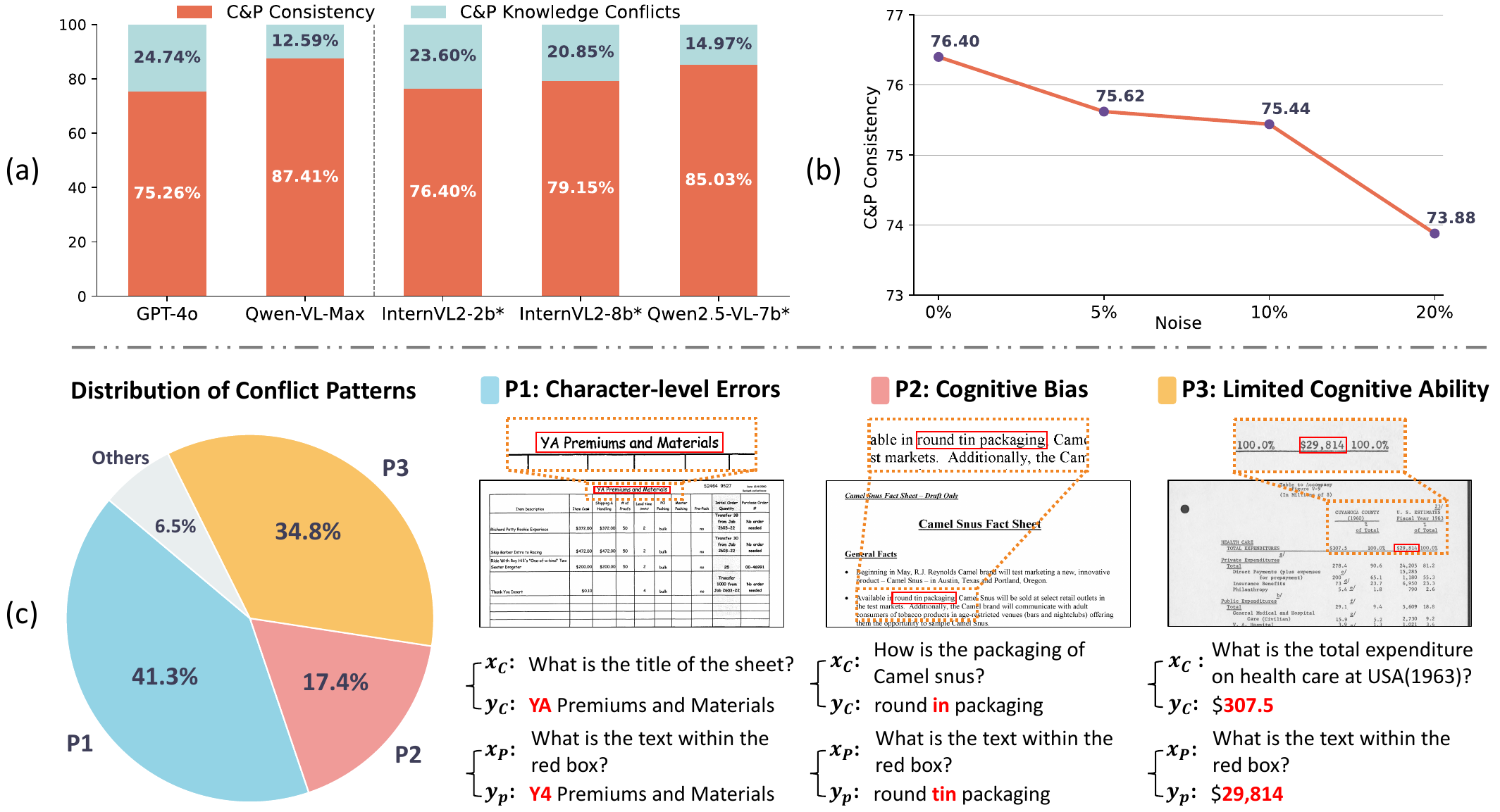} 
    \caption{\textit{a}: C\&P knowledge conflicts in current MLLMs. ``*'' denotes the ``SFT-baseline'' (see Section \ref{sec:models}). Additional quantitative results are provided in Section~\ref{app:mllm_inference_prompt} and Table~\ref{tab:detailed_main_results}. \textit{b}: Results of the synthetic noise experiment, with additional details provided in Section~\ref{app:noise}. \textit{c}: The distribution of conflict patterns, including character-level errors (P1), cognitive bias (P2), and limited cognitive ability (P3), with one illustrative example for each.}
    \label{fig:pattern}
\end{figure*}

\subsection{The Construction of Evaluation Samples}
\label{sec:data_construct}

To calculate C\&P consistency, we construct pairs of cognitive (VQA) query and perceptual (OCR) query, i.e., $(x_C, x_P)$,  with each pair using the same ground truth $GT$ from the image $x_I$. The process is as follows:

Given an image $x_I$ with its QA annotation $(Q, A)$, we assign $GT = A$ and $x_C = Q$. We construct $x_P$ using visual prompting \citep{wu2024visual,yang2023set}. $x_P$ is a simple question: ``\texttt{What is the text within the red box?}'' The corresponding image $x_I^B$ is obtained by drawing a red box in $x_I$ at the location of $Box$, denoted as $x_I^B = \text{VisP}(x_I, Box)$, where $\text{VisP}(\cdot)$ represents the visual prompting process and $Box$ is the bounding box containing $GT$. In practice, responses for cognitive and perceptual tasks are obtained as \( y_C = f(x_C, x_I) \) and \( y_P = f(x_P, x_I^B) \), respectively. 

Additionally, constructing $(x_C, x_P)$ pairs involves several preprocessing steps. According to the definition in Section \ref{sec:definition}, the questions must pertain to the text in the image. However, certain questions, such as those related to comparisons or yes/no answers, do not directly reference the text. Moreover, since the current document datasets do not provide $Box$ annotations, we also need to locate $Box$ based on the OCR annotations of $x_I$. We employ GPT-4o to perform these preprocessing steps. Specific details are provided in Section \ref{app:data_construction}.

In particular, we consider five document understanding datasets to construct evaluation samples, which are categorized into three tasks: Document Question Answering (DocVQA \citep{docvqa} and DUDE \citep{dude}), Document Information Extraction (DeepForm \citep{deepform} and FUNSD \citep{funsd}), and Chart Question Answering (ChartQA \citep{chartqa}). The evaluation samples are constructed from the test sets of these datasets. Section~\ref{sec:dataset} and \ref{app:data_construction} provides additional details, including dataset descriptions, an example evaluation sample, and comprehensive statistics.

\begin{figure*}[t]
    \centering
    \includegraphics[width=0.99\textwidth]{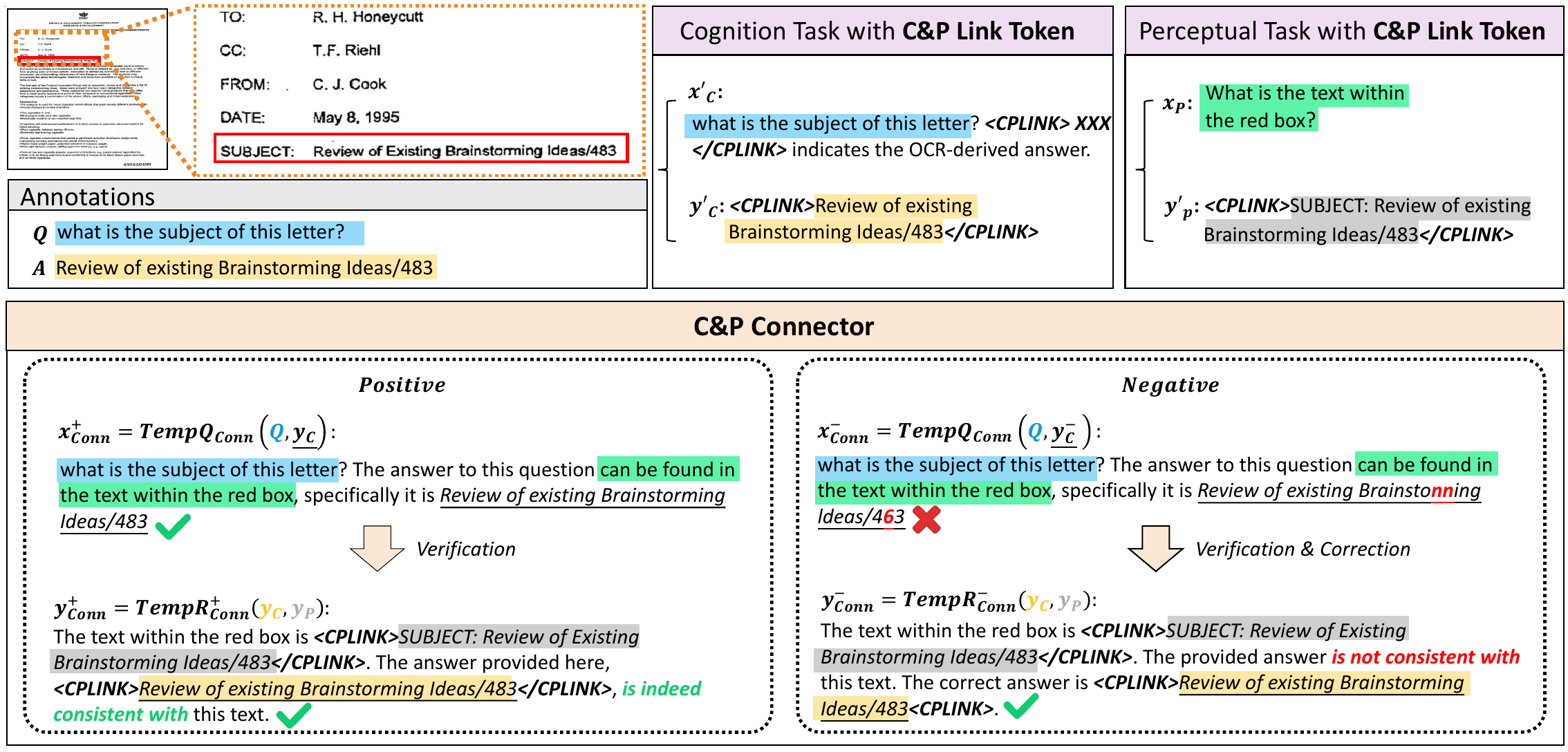} 
    \caption{An example illustrates the source data and its corresponding \textit{Multimodal Knowledge Consistency Fine-tuning} sample.  All mathematical symbols in the figure are consistent with those in Section \ref{sec:method}. Corresponding relationships use the same colors for clarity.}
    \label{fig:method}
\end{figure*}

\section{The Cognition and Perception Knowledge Conflicts in Current MLLMs}
\label{sec:models}
Two closed-source and three open-source MLLMs are evaluated. The closed-source models, GPT-4o \citep{gpt4o} and Qwen-VL-Max \citep{qwen-vl, bai2025qwen2}, are well-regarded in the community. We evaluate these models using their publicly available APIs, disabling all randomness-inducing hyperparameters. Additionally, to ensure that MLLMs follow instructions, we carefully adjust the prompts based on the characteristics of each dataset. Details are provided in Section \ref{app:mllm_inference_prompt}.

The open-source models include InternVL2-2b \citep{internvl}, InternVL2-8b \citep{internvl}, and Qwen2.5-VL-7b \citep{bai2025qwen2}, which differ in size and architecture. We perform the evaluation by disabling all randomness-inducing hyperparameters on an Nvidia A100 GPU. Furthermore, we observe that using the original weights for inference leads to issues with instruction following (see Section \ref{app:mllm_inference_prompt} for details), and thus we construct SFT data using the training sets from all datasets following the procedure outlined in Section \ref{sec:data_construct} to train baseline models for each MLLM, referred to as ``SFT-baseline''. Training details are provided in Section \ref{sec:implement}.

Figure~\ref{fig:pattern} (a) presents the evaluation results, with more quantitative results provided in Section~\ref{app:mllm_inference_prompt} and Table~\ref{tab:detailed_main_results}. Overall, C\&P knowledge conflicts are common in current MLLMs, with inconsistencies observed in 12\%--25\% of cases. Furthermore, the severity of these conflicts appears comparable between open-source and closed-source models.

To further investigate the potential cause of C\&P knowledge conflicts, we train InternVL2-2b with varying levels of synthetic noise (OCR: shape mix-ups, missing or extra letters; VQA: typos, omitted details). Synthetic noise is injected into 5\%, 10\%, and 20\% of the training data. Figure~\ref{fig:pattern} (b) shows the results, with additional quantitative analysis provided in Section~\ref{app:noise}. Overall, as the level of noise increases, the C\&P consistency declines.

We also randomly sample 10\% of all inconsistent cases generated by InternVL2-2b and manually inspect them. Three main types of conflicts are identified, as shown in Figure~\ref{fig:pattern} (c). The majority of conflicts (41.3\%, P1) stem from character-level errors when the model responds to either the OCR or VQA query. P2 (17.4\%) arises from cognitive bias. Although the model ``sees'' the correct text (its OCR output is accurate), it prefers a linguistically more plausible answer (e.g., substituting ``round tin packaging'' with ``round in packaging''). The synthetic noise experiment, together with P1 and P2, supports our hypothesis that heterogeneous VQA and OCR annotations are a primary source of C\&P knowledge conflicts. Specifically, P1 reflects perception noise introduced by external OCR engine annotators, while P2 reflects semantic bias introduced by human or LLM annotators. P3 (34.8\%) reveals a limitation in the model’s cognitive ability, where the VQA response is hallucinated despite an accurate OCR output. To focus on purer conflict conditions, we exclude P3 from the idealized C\&P consistency (Section \ref{sec:definition}).
\section{Multimodal Knowledge Consistency Fine-tuning}
\label{sec:method}

Section \ref{sec:models} demonstrates that even state-of-the-art MLLMs exhibit C\&P knowledge conflicts. To resolve these conflicts, we propose \textit{Multimodal Knowledge Consistency Fine-tuning}, illustrated in Figure \ref{fig:method}, which comprises two components: \textit{C\&P Link Tokens} and the \textit{C\&P Connector}. As heterogeneous VQA and OCR annotations are a primary source of C\&P conflicts (Section \ref{sec:models}), this method aims to reinforce the connection between cognitive and perceptual tasks, thereby mitigating C\&P knowledge conflicts.

\begin{table*}[t]
\begin{center}
\begin{tabular}{lccccc|c}
\toprule
 & DocVQA & DUDE & DeepForm & FUNSD & ChartQA & Average \\
\midrule
InternVL2-2b* & \makecell[c]{80.59\vspace{-0.1cm}\\\scriptsize90.62 } & \makecell[c]{64.69\vspace{-0.1cm}\\\scriptsize83.00 } & \makecell[c]{72.05\vspace{-0.1cm}\\\scriptsize77.40 } & \makecell[c]{80.84\vspace{-0.1cm}\\\scriptsize87.95 } & \makecell[c]{83.80\vspace{-0.1cm}\\\scriptsize91.27 } & \makecell[c]{76.40\vspace{-0.1cm}\\\scriptsize86.05 } \\
InternVL2-2b (Ours) & \makecell[c]{\textbf{83.39}\vspace{-0.1cm}\\\scriptsize91.32 } & \makecell[c]{\textbf{69.49}\vspace{-0.1cm}\\\scriptsize84.75 } & \makecell[c]{\textbf{78.56}\vspace{-0.1cm}\\\scriptsize82.20 } & \makecell[c]{\textbf{81.50}\vspace{-0.1cm}\\\scriptsize89.60 } & \makecell[c]{\textbf{87.64}\vspace{-0.1cm}\\\scriptsize93.17 } & \makecell[c]{\textbf{80.12}\vspace{-0.1cm}\\\scriptsize88.21 } \\
\midrule
InternVL2-8b* & \makecell[c]{84.28\vspace{-0.1cm}\\\scriptsize91.32 } & \makecell[c]{67.82\vspace{-0.1cm}\\\scriptsize83.14 } & \makecell[c]{74.19\vspace{-0.1cm}\\\scriptsize77.70 } & \makecell[c]{82.60\vspace{-0.1cm}\\\scriptsize91.82 } & \makecell[c]{86.88\vspace{-0.1cm}\\\scriptsize91.86 } & \makecell[c]{79.15\vspace{-0.1cm}\\\scriptsize87.17 } \\
InternVL2-8b (Ours) & \makecell[c]{\textbf{87.32}\vspace{-0.1cm}\\\scriptsize93.03 } & \makecell[c]{\textbf{73.26}\vspace{-0.1cm}\\\scriptsize84.70 } & \makecell[c]{\textbf{79.17}\vspace{-0.1cm}\\\scriptsize82.22 } & \makecell[c]{\textbf{83.48}\vspace{-0.1cm}\\\scriptsize90.13 } & \makecell[c]{\textbf{90.53}\vspace{-0.1cm}\\\scriptsize94.22 } & \makecell[c]{\textbf{82.75}\vspace{-0.1cm}\\\scriptsize88.86 } \\
\midrule
Qwen2.5-VL-7b* & \makecell[c]{93.79\vspace{-0.1cm}\\\scriptsize96.44 } & \makecell[c]{79.30\vspace{-0.1cm}\\\scriptsize91.87 } & \makecell[c]{75.20\vspace{-0.1cm}\\\scriptsize85.36 } & \makecell[c]{84.80\vspace{-0.1cm}\\\scriptsize90.38 } & \makecell[c]{92.06\vspace{-0.1cm}\\\scriptsize95.37 } & \makecell[c]{85.03\vspace{-0.1cm}\\\scriptsize91.88 } \\
Qwen2.5-VL-7b (Ours) & \makecell[c]{\textbf{94.95}\vspace{-0.1cm}\\\scriptsize97.10 } & \makecell[c]{\textbf{84.04}\vspace{-0.1cm}\\\scriptsize94.22 } & \makecell[c]{\textbf{79.57}\vspace{-0.1cm}\\\scriptsize86.73 } & \makecell[c]{\textbf{90.31}\vspace{-0.1cm}\\\scriptsize94.07 } & \makecell[c]{\textbf{93.09}\vspace{-0.1cm}\\\scriptsize95.74 } & \makecell[c]{\textbf{88.39}\vspace{-0.1cm}\\\scriptsize93.57 } \\
\bottomrule
\end{tabular}
\end{center}
\caption{Performance comparison between the original MLLM and the MLLM after multimodal knowledge consistency fine-tuning (ours) across all datasets. All values are percentages (\%). The main number is C\&P Consistency, and the smaller number is Idealized C\&P Consistency. Bolded numbers indicate superior performance. The average results are the macro-averages of all datasets. ``*'' denotes the ``SFT-baseline'' (see Section \ref{sec:models}).}
\label{tab:detailed_main_results}
\end{table*}

\subsection{C\&P Link Tokens}

Previous research \citep{wu2024visionllm} indicates that special tokens can effectively connect knowledge across different tasks. Therefore, we define a pair of \textit{C\&P Link Tokens} to connect cognitive and perceptual tasks, namely \texttt{<CPLINK>} and \texttt{</CPLINK>}, and add them to the original MLLM vocabulary. 
When the MLLM responds to a query using text extracted from an image, it encloses that text with the two C\&P link tokens, for example, ``\texttt{<CPLINK>XXX</CPLINK>}.'' Given an image \( x_I \) with QA annotation \( (Q, A) \), the cognitive task’s query and response are \( (x_C, y_C) \) and the perceptual task’s are \( (x_P, y_P) \).
According to Section \ref{sec:data_construct}, both $y_C$ and $y_P$ are texts derived from $x_I$, i.e. $A$. Therefore, the C\&P link tokens can be applied to the responses of both tasks, denoted as $y_C'$ and $y_P'$, thereby strengthening their connection. Additionally, for guiding linked responses,
we design $x_C'$ to more explicitly prompt the model by adding a special instruction: ``\texttt{<CPLINK>XXX</CPLINK> indicates the OCR-derived answer}.''

\subsection{C\&P Connector}
The second component is the \textit{C\&P Connector}, which uses the question \( Q \) as an intermediary to link \( y_P \) and \( y_C \), thereby bridging cognitive and perceptual tasks. The C\&P Connector consists of positive and negative samples, denoted as \( (x_{Conn}^+, y_{Conn}^+) \) and \( (x_{Conn}^-, y_{Conn}^-) \), respectively. In terms of input images, the connector takes images with bounding boxes, \( x_I^B \), as input (see Section \ref{sec:data_construct} for details).

Positive samples aim to guide the model to use perceptual knowledge to verify cognitive knowledge. Specifically, as shown in Figure~\ref{fig:method}, \( (x_{Conn}^+, y_{Conn}^+) \) is constructed as follows:

\begin{equation}
\left\{
\begin{array}{l}
x_{Conn}^+ = \text{TempQ}_{Conn}(Q, y_C) \\
y_{Conn}^+ = \text{TempR}_{Conn}^+(y_C, y_P)
\end{array}.
\right.
\end{equation}

Here, \( \text{TempQ}_{Conn}(\cdot) \) is the template for constructing C\&P connector queries, and \( \text{TempR}_{Conn}^+(\cdot) \) is the template for constructing positive sample responses. The model is required to first answer \( y_P \), and then \( y_C \), thus using perceptual knowledge to verify cognitive knowledge.

In addition to verification, negative samples further guide the model to use perceptual knowledge to correct erroneous cognitive results. Specifically, as shown in Figure~\ref{fig:method}, \( (x_{Conn}^-, y_{Conn}^-) \) is constructed as follows:

\begin{equation}
\left\{
\begin{array}{l}
x_{Conn}^- = \text{TempQ}_{Conn}(Q, y_C^-) \\
y_{Conn}^- = \text{TempR}_{Conn}^-(y_C, y_P)
\end{array}.
\right.
\end{equation}

Here, the template for constructing queries is the same as that used for positive samples. \( y_C^- \) is an OCR-error version of \( y_C \), generated using GPT-4o (refer to the Section \ref{app:distrub} for the specific prompt). \( \text{TempR}_{Conn}^-(\cdot) \) is the template for generating negative sample responses, which require the model to first answer \( y_P \), then indicate that \( y_C^- \) is incorrect, and finally provide the correct \( y_C \).

The final training data, given \( N \) pairs of \( (Q, A) \), is represented as follows:

\begin{equation}
    \begin{aligned}
    \mathcal{X} = \{& (x'_{C_i}, y'_{C_i}),  (x_{P_i}, y'_{P_i}), \\
    & (x_{Conn_i}^+, y_{Conn_i}^+), (x_{Conn_i}^-, y_{Conn_i}^-)\}_{i=0}^{N}.
    \end{aligned}
\end{equation}

\begin{table*}[t]
\begin{center}
\setlength\tabcolsep{4pt}
\begin{tabular}{lcccccccccc}
\toprule
 & \multicolumn{2}{c}{Doc} & \multicolumn{2}{c}{\multirow{2}{*}{DUDE}} & \multicolumn{2}{c}{Deep} & \multicolumn{2}{c}{\multirow{2}{*}{FUNSD}} & \multicolumn{2}{c}{Chart} \\
 & \multicolumn{2}{c}{VQA} & \multicolumn{2}{c}{} & \multicolumn{2}{c}{Form} & \multicolumn{2}{c}{} & \multicolumn{2}{c}{QA} \\
 & C.T. & P.T. & C.T. & P.T. & C.T. & P.T. & C.T. & P.T. & C.T. & P.T. \\
 \midrule
InternVL2-2b* & 83.44 & 91.71 & 60.29 & 86.64 & 72.42 & 91.70 & 73.87 & 87.39 & 72.76 & 96.39 \\
InternVL2-2b (Ours) & \textbf{85.37} & \textbf{93.24} & \textbf{62.44} & \textbf{88.78} & \textbf{75.50} & \textbf{94.09} & \textbf{76.34} & \textbf{88.69} & \textbf{75.84} & \textbf{97.28} \\
\midrule
InternVL2-8b* & 88.54 & 92.27 & 65.09 & 88.88 & 76.58 & 92.70 & 78.01 & 87.33 & 78.52 & 96.95 \\
InternVL2-8b (Ours) & \textbf{89.47} & \textbf{94.01} & \textbf{67.18} & \textbf{90.41} & \textbf{77.08} & \textbf{94.58} & \textbf{78.16} & \textbf{89.77} & \textbf{82.80} & \textbf{97.55} \\
\midrule
Qwen2.5-VL-7b* & 94.79 & 90.67 & 70.11 & 87.56 & 50.17 & 95.64 & 79.75 & 89.39 & 87.76 & 95.29 \\
Qwen2.5-VL-7b (Ours) & \textbf{95.40} & \textbf{91.85} & \textbf{71.10} & \textbf{88.66} & \textbf{57.58} & \textbf{96.90} & \textbf{80.52} & \textbf{91.29} & \textbf{88.32} & \textbf{95.74} \\
\bottomrule
\end{tabular}
\end{center}
\caption{The performance of cognitive and perceptual tasks. ``C.T.'' and ``P.T.'' stand for cognitive task (VQA) and perceptual task (OCR), respectively. Metrics are detailed in Section~\ref{sec:performance_c_p}; all values are percentages (\%), with bold indicating superior performance.  ``*'' denotes the ``SFT-baseline'' (see Section \ref{sec:models}).}
\label{tab:performance_c_p}
\end{table*}

\section{Experiment}
\subsection{Implementation}
\label{sec:implement}
We construct the training data using the training sets from the five datasets mentioned in Section \ref{sec:data_construct}. For the multimodal knowledge consistency fine-tuning experiment, we focus on three open-source MLLMs (Section \ref{sec:models}): InternVL2-2b, InternVL2-8b, and Qwen2.5-VL-7b. We train all models using the original weights from Huggingface with a learning rate of 1e-5 and a batch size of 128, while keeping other hyperparameters at their default settings. We freeze the visual encoder and optimize only the language model. Each model trains for 1 epoch using 8 Nvidia A100 GPUs. We disable all randomness-inducing hyperparameters during inference.

\subsection{C\&P Consistency Results}
\label{sec:main_results}
The evaluation is conducted on the dataset constructed in Section \ref{sec:data_construct}. The experimental results, presented in Table \ref{tab:detailed_main_results}, demonstrate that our multimodal knowledge consistency fine-tuning method enhances C\&P consistency across all five datasets.
Specifically, InternVL2-2b and InternVL2-8b show improvements of 3.72\% and 3.60\% in C\&P consistency, respectively, while Qwen2.5-VL-7b exhibits a 3.36\% increase. Under ideal conditions, consistency also improves. These findings indicate that our method effectively reduces C\&P knowledge conflicts by linking perceptual and cognitive tasks. 
The comparison between Qwen2.5-VL-7b and the InternVL2 models highlights the general applicability of our approach across different MLLM architectures. 
Additionally, we perform two-sided paired t-tests using InternVL2-2b in Section~\ref{app:t-test}, showing that all gains in Table~\ref{tab:detailed_main_results} are statistically significant.

\begin{table}[t]
\begin{center}
\setlength\tabcolsep{3.5pt}
\begin{tabular}{lcccccc|c}
    \toprule
    \# & Link.      & Conn.      &  & 
    \makecell[c]{Doc\\VQA}   & 
    \makecell[c]{Deep\\Form} & 
    \makecell[c]{Chart\\QA}  & 
    Average                       \\
    \midrule
\midrule
1 &            &            &  & \makecell[c]{80.59\vspace{-0.1cm}\\\scriptsize90.62 } & \makecell[c]{72.05\vspace{-0.1cm}\\\scriptsize77.40 } & \makecell[c]{83.80\vspace{-0.1cm}\\\scriptsize91.27 } & \makecell[c]{76.40\vspace{-0.1cm}\\\scriptsize86.05 } \\
\midrule
2 &            & \checkmark &  & \makecell[c]{82.97\vspace{-0.1cm}\\\scriptsize91.52 } & \makecell[c]{77.85\vspace{-0.1cm}\\\scriptsize80.97 } & \makecell[c]{87.45\vspace{-0.1cm}\\\scriptsize93.66 } & \makecell[c]{79.10\vspace{-0.1cm}\\\scriptsize87.77 } \\
3 & \checkmark &            &  & \makecell[c]{82.71\vspace{-0.1cm}\\\scriptsize91.14 } & \makecell[c]{77.24\vspace{-0.1cm}\\\scriptsize80.72 } & \makecell[c]{87.45\vspace{-0.1cm}\\\scriptsize93.26 } & \makecell[c]{79.24\vspace{-0.1cm}\\\scriptsize87.51 } \\
\midrule
4 & \checkmark & \checkmark &  & \makecell[c]{\textbf{83.39}\vspace{-0.1cm}\\\scriptsize91.32 } & \makecell[c]{\textbf{78.56}\vspace{-0.1cm}\\\scriptsize82.20 } & \makecell[c]{\textbf{87.64}\vspace{-0.1cm}\\\scriptsize93.17 } & \makecell[c]{\textbf{80.12}\vspace{-0.1cm}\\\scriptsize88.21 } \\
\bottomrule
\end{tabular}
\end{center}
\caption{Ablation study based on InternVL2-2b. All values are percentages (\%), with the primary number representing C\&P Consistency and the smaller representing Idealized C\&P Consistency. The best results are in bold. ``Link.'' and ``Conn.'' denote C\&P link token and C\&P connector, respectively (see Section~\ref{sec:method}).}
\label{tab:ablation_study_con}
\end{table}

\begin{table}[t]
\begin{center}
\setlength\tabcolsep{2.5pt}
\begin{tabular}{lcccccccc}
\toprule
  &            &            & \multicolumn{2}{c}{Doc} & \multicolumn{2}{c}{Deep} & \multicolumn{2}{c}{Chart} \\
  &            &            & \multicolumn{2}{c}{VQA} & \multicolumn{2}{c}{Form} & \multicolumn{2}{c}{QA}    \\
\# & Link.      & Conn.      & C.T.       & P.T.       & C.T.        & P.T.       & C.T.        & P.T.        \\
\midrule
1 &            &            & 83.4       & 91.7       & 72.4        & 91.7       & 72.8        & 96.4        \\
\midrule
2 &            & \checkmark & 85.0       & 92.9       & 75.3        & 93.5       & 75.6        & 96.8        \\
3 & \checkmark &            & 85.1       & 93.1      & 75.2        & 94.0       & 75.4        & 97.1        \\
\midrule
4 & \checkmark & \checkmark & \textbf{85.4}       & \textbf{93.2}       & \textbf{75.5}        & \textbf{94.1}       & \textbf{75.8 }       & \textbf{97.3}        \\
\bottomrule
\end{tabular}
\end{center}
\caption{Ablation study based on InternVL2-2b.  ``C.T.'' and ``P.T.'' denote cognitive (VQA) and perceptual (OCR) tasks. Metrics are in Section~\ref{sec:performance_c_p}; values are percentages (\%), with bold numbers indicating best performance. ``Link.'' and ``Conn.'' denote C\&P link token and C\&P connector, respectively (see Section~\ref{sec:method}).}
\label{tab:ablation_study_performance}
\end{table}



\subsection{The Performance of Cognitive and Perceptual Tasks}
\label{sec:performance_c_p}
To assess the impact of C\&P consistency on model performance, we evaluate the model's effectiveness on cognitive and perceptual tasks.
For the cognitive task, following previous works \citep{borchmann2021due,lee2023pix2struct,layoutllm}, we evaluate DocVQA and FUNSD using ANLS \citep{biten2019scene}, DeepForm using the F1 score, and ChartQA using relaxed accuracy \citep{methani2020plotqa}. For the perceptual task, all datasets are evaluated using ANLS.

As shown in Table \ref{tab:performance_c_p}, the three MLLMs show improved performance on both cognitive and perceptual tasks across all datasets after the multimodal knowledge consistency fine-tuning.
We attribute this improvement to our fine-tuning approach, which reduces the conflict between perceptual and cognitive knowledge, thereby promoting their integration. We believe that the results suggest that enhancing C\&P consistency can strengthen the capabilities of MLLMs.
Similar to Section~\ref{sec:main_results}, the t-tests in Section~\ref{app:t-test} show that the performance gains are statistically significant.

\subsection{Ablation Study}
\label{sec:ablation}

To evaluate the contribution of each component in multimodal knowledge consistency fine-tuning, we conduct a series of ablation experiments using InternVL2-2b, as shown in Table \ref{tab:ablation_study_con} and Table \ref{tab:ablation_study_performance}. Due to space limits, we show three datasets here and provide the rest in Section~\ref{app:ablation}. Each experiment, with different fine-tuning tasks, is trained according to the settings outlined in Section \ref{sec:implement}. \#2 removes all C\&P link tokens from the training data, including those in the C\&P connector. The results in Table \ref{tab:ablation_study_con} validate our hypothesis that both components in multimodal knowledge consistency fine-tuning are crucial for enhancing C\&P consistency. For instance, on average, the C\&P link token improves by 2.70\%, and the C\&P connector improves by 2.84\%. Furthermore, Table \ref{tab:ablation_study_performance} shows that our method achieves the best performance on cognitive and perceptual tasks.

\begin{figure}[t]
    \centering
    \includegraphics[width=1 \columnwidth]{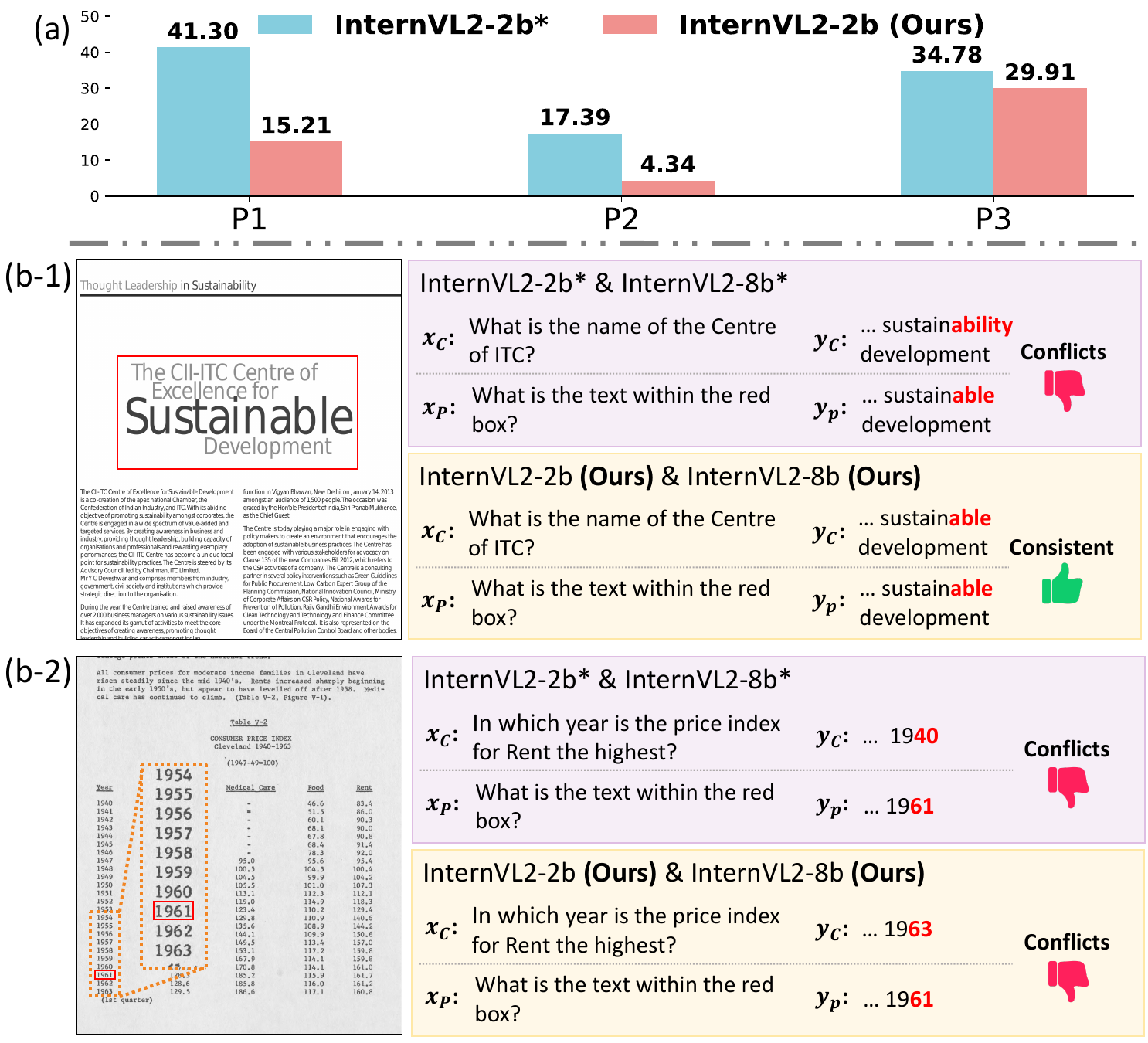} 
    \caption{\textit{a}: Comparison of the distribution of conflict patterns between InternVL2-2b* and InternVL2-2b (Ours). \textit{b}: Two cases: \textit{b-1} demonstrates the effectiveness of our method, while \textit{b-2} reveals a limitation.}
    \label{fig:case_study}
\end{figure}

\subsection{Analysis of Conflict Patterns and Case Evidence}

To further evaluate the effectiveness of multimodal knowledge consistency fine-tuning, we reuse the procedure described in Section~\ref{sec:models} to analyze conflict patterns. Figure~\ref{fig:case_study} (a) shows that, after fine-tuning, character-level errors (P1) and cognitive bias (P2) decrease significantly, making limited cognitive ability (P3) the dominant pattern. This shift supports our claim that heterogeneous VQA and OCR annotations are the primary sources of C\&P knowledge conflicts and confirms that our method mitigates them effectively. The qualitative evidence in Figure~\ref{fig:case_study} (b) illustrates these statistics. In case (b-1), categorized as P2, both InternVL2-2b and InternVL2-8b recognize ``sustainable development'' but incorrectly respond with ``sustainability development'' in the VQA task due to cognitive bias. The conflicts disappear after fine-tuning, as the models better integrate cognitive and perceptual knowledge. Notably, a similar P1 case with the same conclusion is provided in the Section \ref{app:case_study}. In case (b-2), categorized as P3, the result indicates that our method cannot fundamentally extend the model’s cognitive boundaries. In Figure~\ref{fig:case_study}, the responses of InternVL2-2b and InternVL2-8b are identical, reflecting the representativeness of these cases, though they differ in most other cases.

\section{Related Work}

\paragraph{MLLMs for Document Understanding}
Document understanding \citep{cui2021document,xu2021layoutlmv2,huang2022layoutlmv3,luo2023geolayoutlm,layoutllm,shao2023gem, wang2023docllm,Zhu_2025_CVPR,mo2025doc} is a rapidly growing research area driven by increasing industrial demand. Its main objective is to comprehend complex typeset images that contain rich textual information, such as scanned document pages \citep{docvqa,deepform,klc}, charts \citep{chartqa,kafle2018dvqa,methani2020plotqa}, tables \citep{wtq,chen2019tabfact, mo2025tablenarrator}, and other formats \citep{tanaka2021visualmrc,mathew2022infographicvqa,xing2024dochienet, shao2024webrpg}. As a multimodal task, document understanding involves automated processes for understanding, classifying, and extracting information, requiring models to possess both perceptual and cognitive capabilities \citep{cui2021document}.
Recent studies \citep{internvl,hong2024cogagent,dong2024internlm, bai2025qwen2} for general MLLMs improve the encoding resolution of document images, significantly boosting performance in document understanding tasks.
Several MLLMs are developed to focus on addressing document understanding problems, such as mPLUG-DocOwl \citep{ye2023mplugdocowl,docowl15} and UReader \citep{ye2023ureader}.

\paragraph{Knowledge Conflicts in LLMs}

LLMs are distinguished for encapsulating an extensive repository of world knowledge, known as the memory. Simultaneously, LLMs continue to engage with external contextual knowledge post-deployment \citep{pan2023knowledge}. The discrepancies between the contexts and the model’s memory knowledge, i.e. context-memory conflicts, are being intensively studied recently \citep{xie2023adaptive}. Another notable challenge arises with intra-memory conflict—a condition where LLMs exhibit unpredictable behaviors to inputs that are semantically equivalent but syntactically distinct \citep{chang2023languagemodelbehaviorcomprehensive,bartsch2023self,gvconsistency, zhu2024unraveling, zhang2024cross}. This variance can be attributed to the conflicting knowledge embedded within the LLM’s memory, which stems from the inconsistencies present in the complex and diverse pre-training datasets. 

\paragraph{Hallucination issues in MLLMs}

MLLMs provide powerful tools for content generation across a wide range of tasks.
However, they are susceptible to hallucinations \citep{bang-etal-2023-multitask,zhang2023sirenssongaiocean, liu2024insight}, where the generated outputs contain information not present in the visual input. These hallucinations typically arise when the models overly rely on the strong priors of their language modules. Such conflicts between MLLMs' language and visual perception raise concerns about their reliability and limit their applications \citep{Ji_2023,kaddour2023challenges}. Current research primarily focuses on detecting and evaluating hallucinations \citep{li2023evaluating,zhang2023sirenssongaiocean}, as well as methods to reduce them \citep{liu2024mitigating,hevigc2024}. To mitigate hallucinations, efforts have been directed toward enhancing data collection and training procedures \citep{liu2024mitigating,hevigc2024}. Nevertheless, research on how MLLMs integrate perception and cognition knowledge, which is also vital for interpreting and debugging these models, has not progressed at the same pace. 
\section{Conclusion}

In this paper, we identify that current MLLMs often face conflicts between cognition and perception, referred to as \textit{Cognition and Perception (C\&P) knowledge conflicts}.
The severity of these conflicts is systematically assessed across five document understanding datasets, revealing that even leading MLLMs still struggle with these multimodal knowledge conflicts.
To address this problem, a novel method called \textit{Multimodal Knowledge Consistency Fine-tuning} is introduced. 
Comprehensive experiments demonstrate the effectiveness of our method in reducing C\&P knowledge conflicts. 
Additionally, our method improves the performance of MLLMs in both cognitive and perceptual tasks. 


\section*{Limitations}
Despite contributing to the identification and mitigation of C\&P knowledge conflicts, several limitations remain. This work simplifies cognition and perception to VQA and OCR tasks, potentially overlooking other cognitive abilities (e.g., multi-step reasoning, layout-aware inference) and perceptual channels (e.g., color, shape). We address these omissions in future work. Moreover, the current focus is on document understanding. We plan to extend our research to broader multimodal domains, such as general open-world images and video streams, to further explore C\&P knowledge conflicts.

\section*{Acknowledgments}
This work is supported by the National Natural Science Foundation of China (Grant No. 62372408).


\bibliography{custom}

\appendix

\clearpage

\section{Additional Details}
\subsection{Details of Selected Datasets}
\label{sec:dataset}

We consider five document understanding datasets to assess C\&P knowledge conflicts, categorized into the following three tasks:

\textbf{Document QA.} DocVQA \citep{docvqa} contains 50k question-answer pairs from 12k document images in the UCSF Industry Documents Library. DUDE \citep{dude} covers diverse domains, including medical, legal, technical, and financial, providing 41k question-answer pairs from 5k documents. We exclude all multi-page VQA annotations from DUDE, retaining only single-page annotations.

\textbf{Document IE.} DeepForm \citep{deepform} and FUNSD \citep{funsd} are two Information Extraction datasets. DeepForm consists of 1.1k documents related to election spending. FUNSD contains 0.2k document images from the RVL-CDIP dataset \citep{harley2015evaluation}. The annotations for DeepForm and FUNSD are transformed into a question-answer format, with DeepForm following  \citet{docowl15}, and FUNSD following  \citet{layoutllm}. The annotations in  \citet{docowl15} for DeepForm incorrectly assume that all key values are on the first page, ignoring that DeepForm documents are multi-page. We correct this issue (see Section \ref{app:deepform} for details), ensuring information extraction occurs on the correct pages.

\textbf{Chart QA.} ChartQA \citep{chartqa} compiles a diverse range of topics and chart types from four primary sources: Statista (statista.com), The Pew Research Center (pewresearch.org), Our World in Data (ourworldindata.org), and the OECD (oecd.org). In total, the dataset includes 21k chart images and 32k question-answer pairs.

Notably, OCR annotations are required in Section \ref{sec:data_construct}. For DocVQA and DUDE, the official OCR annotations are utilized, whereas the other datasets employ OCR annotations generated by a commercial OCR solution.

\subsection{Details of Evaluation Sample Construction}
\label{app:data_construction}
\begin{figure}[t]
    \centering
    \includegraphics[width=0.99\columnwidth]{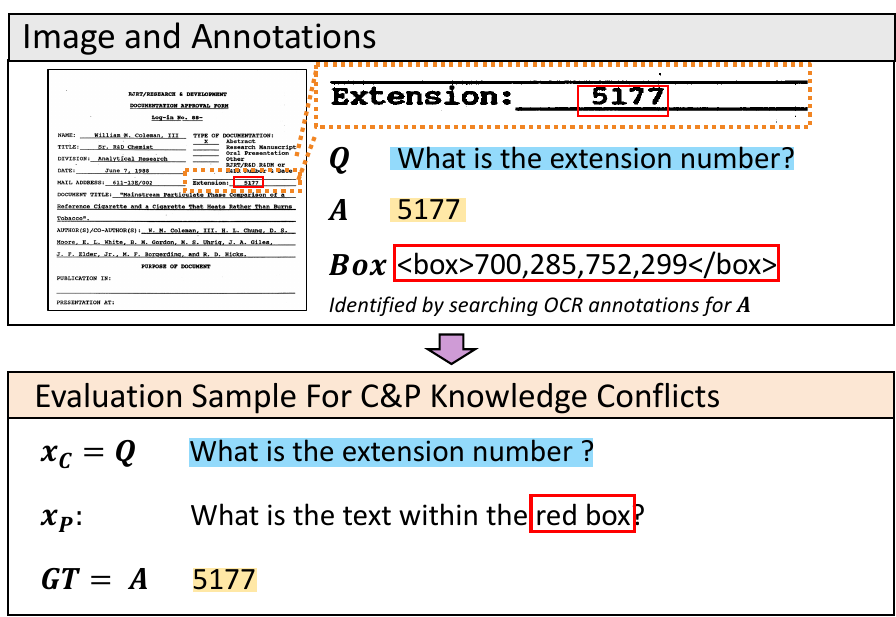} 
    \caption{A specific example illustrates the evaluation sample. All mathematical symbols in the figure are consistent with those in Section \ref{sec:data_construct}. Corresponding relationships are represented using the same colors for clarity.}
    \label{fig:construct_data}
\end{figure}

As described in Section \ref{sec:data_construct}, the construction of $(x_C, x_P)$ pairs involves several preprocessing steps. According to the definition in Section \ref{sec:definition}, the questions must pertain to the text in the image. However, certain questions, such as those related to comparisons or yes/no answers, do not directly reference the text. To address this, we filter out such QA pairs using GPT-4o with the prompt detailed in Table \ref{tab:prompt_4o_qa_filter}. Moreover, since the $Box$ annotations are not provided, we employ GPT-4o to locate $Box$ based on the OCR annotations of $x_I$ with the prompt detailed in Table \ref{tab:prompt_4o_locate}. We use GPT-4o to find $Box$ because a single image may contain multiple occurrences of the text $A$ in different locations. Therefore, identifying the correct $Box$ requires semantic understanding, which GPT-4o excels at. QA pairs for which GPT-4o cannot find a $Box$, or the $Box$ found does not contain $A$, are also excluded. 
Additionally, an example of an evaluation sample is provided in Section \ref{fig:construct_data}. Table \ref{tab:eval_data_count} provides the statistics of evaluation data, including the number of $(x_C, x_P)$ pairs and their corresponding images.

\begin{table}[t]
\begin{center}
\setlength\tabcolsep{3pt}
\begin{tabular}{lccccc}
\toprule
 & \makecell[c]{Doc\\VQA} & \makecell[c]{DUDE} & \makecell[c]{Deep\\Form} & \makecell[c]{FUNSD} & \makecell[c]{Chart\\QA} \\
\midrule
\# $(x_C,x_P)$  & 4575   & 1855 & 984      & 454   & 1562    \\
\# Images                & 1268   & 1101 & 248      & 46    & 1278    \\
\bottomrule
\end{tabular}
\end{center}
\caption{Data statistics for C\&P knowledge conflicts evaluation. The number of evaluation samples, i.e., cognitive (VQA) query and perceptual (OCR) query \((x_C, x_P)\) pairs, along with the corresponding images for each dataset.}
\label{tab:eval_data_count}
\end{table}

\begin{table*}[t]
\label{tab:inference_results}
\begin{center}
\begin{tabular}{lccccc|c}
\toprule
 & DocVQA & DUDE & DeepForm & FUNSD & ChartQA & Average \\
\midrule
GPT-4o & \makecell[c]{85.58\vspace{-0.1cm}\\\scriptsize93.35 } & \makecell[c]{67.84\vspace{-0.1cm}\\\scriptsize87.30 } & \makecell[c]{62.70\vspace{-0.1cm}\\\scriptsize71.20 } & \makecell[c]{78.76\vspace{-0.1cm}\\\scriptsize90.52 } & \makecell[c]{81.41\vspace{-0.1cm}\\\scriptsize92.38 } & \makecell[c]{75.26\vspace{-0.1cm}\\\scriptsize86.95 } \\
Qwen-VL-Max & \makecell[c]{95.66\vspace{-0.1cm}\\\scriptsize97.15 } & \makecell[c]{82.54\vspace{-0.1cm}\\\scriptsize91.35 } & \makecell[c]{83.23\vspace{-0.1cm}\\\scriptsize86.69 } & \makecell[c]{83.19\vspace{-0.1cm}\\\scriptsize90.52 } & \makecell[c]{92.44\vspace{-0.1cm}\\\scriptsize95.68 } & \makecell[c]{87.41\vspace{-0.1cm}\\\scriptsize92.28 } \\
\bottomrule
\end{tabular}
\caption{C\&P knowledge conflicts in current MLLMs. All values are percentages (\%), where the primary number represents C\&P Consistency and the smaller number represents Idealized C\&P Consistency. }
\label{tab:inference_results_2}
\end{center}
\end{table*}

\begin{table}[t]
\begin{center}
\setlength\tabcolsep{8pt}
\begin{tabular}{lcccc}
\toprule
\# & Link.      & Conn.      & DUDE                                                  & FUNSD                                                 \\
\midrule
1 &            &            & \makecell[c]{64.69\vspace{-0.1cm}\\\scriptsize83.00 } & \makecell[c]{80.84\vspace{-0.1cm}\\\scriptsize87.95 } \\
\midrule
2 &            & \checkmark & \makecell[c]{67.49\vspace{-0.1cm}\\\scriptsize84.14 } & \makecell[c]{79.74\vspace{-0.1cm}\\\scriptsize88.56 } \\
3 & \checkmark &            & \makecell[c]{68.84\vspace{-0.1cm}\\\scriptsize84.46 } & \makecell[c]{79.96\vspace{-0.1cm}\\\scriptsize87.97 } \\
\midrule
4 & \checkmark & \checkmark & \makecell[c]{\textbf{69.49}\vspace{-0.1cm}\\\scriptsize84.75 } & \makecell[c]{\textbf{81.50}\vspace{-0.1cm}\\\scriptsize89.60 } \\
\bottomrule
\end{tabular}
\end{center}
\caption{Ablation study based on InternVL2-2b. All values are percentages (\%), with the primary number representing C\&P consistency and the smaller representing idealized C\&P consistency. Best results are in bold. ``Link.'' and ``Conn.'' denote C\&P link token and C\&P connector, respectively, as detailed in Section \ref{sec:method}.}
\label{tab:ablation_study_con_2}
\end{table}

\begin{table}[t]
\begin{center}
\setlength\tabcolsep{4.5pt}
\begin{tabular}{lcccccc}
\toprule
  &            &            & \multicolumn{2}{c}{\multirow{2}{*}{DUDE}} & \multicolumn{2}{c}{\multirow{2}{*}{FUNSD}} \\
  &            &            & \multicolumn{2}{c}{}                      & \multicolumn{2}{c}{}                       \\
\# & Link.      & Conn.      & C.T.                & P.T.                & C.T.                 & P.T.                \\
\midrule
1 &            &            & 60.29               & 86.64               & 73.87                & 87.39               \\
\midrule
2 &            & \checkmark & 62.32               & 87.56               & 75.89                & 88.02               \\
3 & \checkmark &            & 61.68               & 88.43               & 76.20                & 88.51               \\
\midrule
4 & \checkmark & \checkmark & \textbf{62.44}               & \textbf{88.78}               & \textbf{76.34}                & \textbf{88.69}               \\
\bottomrule
\end{tabular}
\end{center}
\caption{Ablation study based on InternVL2-2b.  ``C.T.'' and ``P.T.'' denote cognitive (VQA) and perceptual (OCR) tasks. Metrics are in Section~\ref{sec:performance_c_p}; values are percentages (\%), with bold numbers indicating best performance. ``Link.'' and ``Conn.'' denote C\&P link token and C\&P connector, respectively (see Section~\ref{sec:method}).}
\label{tab:ablation_study_performance_2}
\end{table}

\subsection{Details of DeepForm Single-page QA Annotations}
\label{app:deepform}
As described in Section \ref{sec:dataset}, \citet{docowl15} provide incorrect annotations for DeepForm because they assume all key values are on the first page, overlooking that DeepForm documents are multi-page. To address this, we use GPT-4o to identify the correct page for information extraction using the prompt detailed in Table \ref{tab:prompt_4o_selection}, ensuring that all single-page QA annotations in DeepForm are correct.

\subsection{Additional Details of C\&P Knowledge Conflicts Evaluation}
\label{app:mllm_inference_prompt}
As described in Section \ref{sec:models}, to ensure that closed-source MLLMs follow instructions, we carefully adjust the prompts based on the characteristics of each dataset. For cognitive tasks, the prompts for DocVQA and DUDE are detailed in Table \ref{tab:prompt_4o_docvqa_dude}, DeepForm in Table \ref{tab:prompt_4o_deepform}, FUNSD in Table \ref{tab:prompt_4o_funsd}, and ChartQA in Table \ref{tab:prompt_4o_cahrtqa}. For perceptual tasks, the prompts are detailed in Table \ref{tab:prompt_4o_ocr}. Table~\ref{tab:inference_results_2} presents the additional evaluation results of C\&P knowledge conflicts in closed-source MLLMs.

Additionally, Table \ref{tab:c_p_performance_2} presents the performance of closed-source MLLMs on cognitive and perceptual tasks. The results demonstrate that closed-source MLLMs perform well on both tasks, indicating that they effectively follow instructions and validating the results reported in Section \ref{sec:models}.

We also report in Table \ref{tab:c_p_performance_2} the performance of open-source MLLMs with original weights on cognitive and perceptual tasks. The results show that open-source MLLMs perform exceptionally poorly on some datasets, highlighting the necessity of using the ``SFT-baseline'' in Section \ref{sec:models}.

We evaluate the closed-source MLLMs via their publicly available APIs. Specifically, we use the snapshots of GPT-4o\footnote{\url{https://platform.openai.com/docs/models/gpt-4o}} from 2024-11-20 and Qwen-VL-Max\footnote{\url{https://www.alibabacloud.com/help/en/model-studio}} from 2025-04-08.  
For open-source MLLMs, we use the model weights available on Hugging Face, including InternVL2-2B\footnote{\url{https://huggingface.co/OpenGVLab/InternVL2-2B}}, InternVL2-8B\footnote{\url{https://huggingface.co/OpenGVLab/InternVL2-8B}}, and Qwen2.5-VL-7B\footnote{\url{https://huggingface.co/Qwen/Qwen2.5-VL-7B-Instruct}}.

\begin{figure}[t]
    \centering
    \includegraphics[width=0.99\columnwidth]{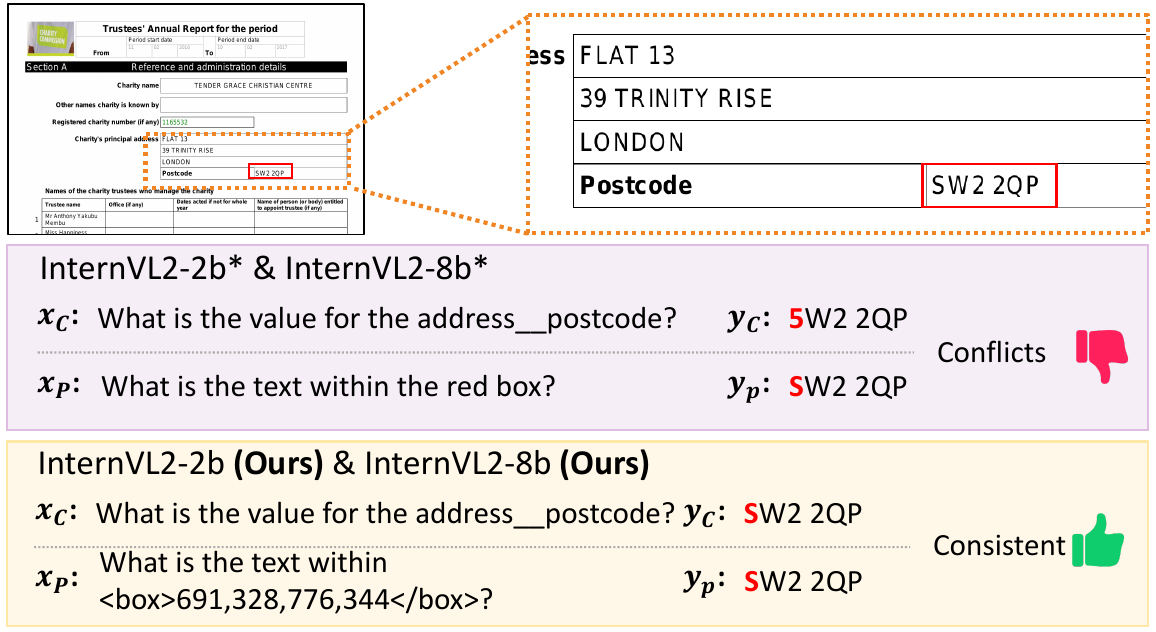} 
    \caption{An additional case demonstrating the effectiveness of our method.}
    \label{fig:case_study2}
\end{figure}

\subsection{Additional Results of the Synthetic Noise Experiment}
\label{app:noise}
The additional results of the synthetic noise experiment based on InternVL2-2b (Section \ref{sec:models}) show in Table \ref{tab:noise}.

\begin{table*}
\begin{center}
\begin{tabular}{lcccccccccc}
\toprule
 & \multicolumn{2}{c}{Doc} & \multicolumn{2}{c}{\multirow{2}{*}{DUDE}} & \multicolumn{2}{c}{Deep} & \multicolumn{2}{c}{\multirow{2}{*}{FUNSD}} & \multicolumn{2}{c}{Chart} \\
 & \multicolumn{2}{c}{VQA} & \multicolumn{2}{c}{} & \multicolumn{2}{c}{Form} & \multicolumn{2}{c}{} & \multicolumn{2}{c}{QA} \\
 & C.T. & P.T. & C.T. & P.T. & C.T. & P.T. & C.T. & P.T. & C.T. & P.T. \\
\midrule
GPT-4o & 89.14 & 86.55 & 62.70 & 74.58 & 37.17 & 85.51 & 75.95 & 87.39 & 68.04 & 95.22 \\
Qwen-VL-Max & 95.88 & 92.25 & 70.94 & 87.91 & 45.33 & 97.00 & 83.18 & 91.06 & 87.48 & 94.28 \\
\midrule
InternVL2-2b & 87.03 & 66.66 & 59.96 & 52.25 & 19.37 & 34.28 & 74.02 & 59.08 & 76.40 & 65.96 \\
InternVL2-8b & 91.73 & 74.18 & 65.96 & 59.80 & 21.63 & 63.67 & 75.84 & 70.22 & 83.12 & 73.05 \\
Qwen2.5-VL-7b & 95.55 & 87.95 & 69.79 & 82.60 & 37.98 & 94.79 & 78.37 & 82.09 & 87.60 & 92.66 \\
\bottomrule
\end{tabular}
\end{center}
\caption{The performance of cognitive and perceptual tasks, consisting of two groups: the results of closed-source models and the results of open-source models with original weights. “C.T.” and “P.T.” stand for cognitive task (VQA) and perceptual task (OCR), respectively. Metrics are detailed in Section \ref{sec:performance_c_p}, and all values are reported as percentages (\%).}
\label{tab:c_p_performance_2}
\end{table*}

\subsection{Additional Results of the Ablation Study}
\label{app:ablation}
Due to space constraints, we report the results of only three datasets in Section \ref{sec:ablation}. The results of the remaining two datasets are presented in Table \ref{tab:ablation_study_con_2} and Table \ref{tab:ablation_study_performance_2}.

\subsection{Additional Details of C\&P Connector}
\label{app:distrub}
As described in Section \ref{sec:method}, the negative samples for the C\&P Connector are required to use the OCR-error version of \( y_C \), denoted as \( y_C^- \), which is generated using GPT-4o with the prompt detailed in Table \ref{tab:prompt_4o_distrub}.


\begin{table*}
\begin{center}
\setlength\tabcolsep{10pt}
\begin{tabular}{lccccc|c}
\toprule
 & DocVQA & DUDE & DeepForm & FUNSD & ChartQA & Average \\
\midrule
0\% & \makecell[c]{\textbf{80.59}\vspace{-0.1cm}\\\scriptsize90.62 } 
   & \makecell[c]{\textbf{64.69}\vspace{-0.1cm}\\\scriptsize83.00 } 
   & \makecell[c]{\textbf{72.05}\vspace{-0.1cm}\\\scriptsize77.40 } 
   & \makecell[c]{\textbf{80.84}\vspace{-0.1cm}\\\scriptsize87.95 } 
   & \makecell[c]{\textbf{83.80}\vspace{-0.1cm}\\\scriptsize91.27 } 
   & \makecell[c]{\textbf{76.40}\vspace{-0.1cm}\\\scriptsize86.05 } \\
5\% & \makecell[c]{79.12\vspace{-0.1cm}\\\scriptsize90.97 } 
   & \makecell[c]{64.09\vspace{-0.1cm}\\\scriptsize82.70 } 
   & \makecell[c]{71.36\vspace{-0.1cm}\\\scriptsize76.31 } 
   & \makecell[c]{79.74\vspace{-0.1cm}\\\scriptsize87.67 } 
   & \makecell[c]{83.81\vspace{-0.1cm}\\\scriptsize90.12 } 
   & \makecell[c]{75.62\vspace{-0.1cm}\\\scriptsize85.55 } \\
10\% & \makecell[c]{79.01\vspace{-0.1cm}\\\scriptsize90.08 } 
    & \makecell[c]{62.58\vspace{-0.1cm}\\\scriptsize81.80 } 
    & \makecell[c]{72.17\vspace{-0.1cm}\\\scriptsize76.85 } 
    & \makecell[c]{80.18\vspace{-0.1cm}\\\scriptsize87.34 } 
    & \makecell[c]{83.24\vspace{-0.1cm}\\\scriptsize90.39 } 
    & \makecell[c]{75.44\vspace{-0.1cm}\\\scriptsize85.29 } \\
20\% & \makecell[c]{77.92\vspace{-0.1cm}\\\scriptsize89.93 } 
    & \makecell[c]{61.61\vspace{-0.1cm}\\\scriptsize81.17 } 
    & \makecell[c]{70.24\vspace{-0.1cm}\\\scriptsize75.38 } 
    & \makecell[c]{77.53\vspace{-0.1cm}\\\scriptsize85.45 } 
    & \makecell[c]{82.08\vspace{-0.1cm}\\\scriptsize90.52 } 
    & \makecell[c]{73.88\vspace{-0.1cm}\\\scriptsize84.49 } \\
\bottomrule
\end{tabular}
\end{center}
\caption{The synthetic noise experiment based on InternVL2-2b. All values are percentages (\%), where the primary number represents C\&P Consistency and the smaller number represents Idealized C\&P Consistency.}
\label{tab:noise}
\end{table*}

\subsection{Additional Case Study}
\label{app:case_study}
We present a case in Figure~\ref{fig:case_study2}, categorized as P1 (Section~\ref{sec:models}), which provides evidence that multimodal knowledge consistency fine-tuning mitigates C\&P knowledge conflicts.

\subsection{Additional Results of the T-test}
\label{app:t-test}
We perform two-sided paired t-tests using InternVL2-2b, and the results are shown in Table \ref{tab:t-test}.

\begin{table*}
\begin{center}
\begin{tabular}{lccccc}
\toprule
 & \textbf{DocVQA} & \textbf{DUDE} & \textbf{DeepForm} & \textbf{FUNSD} & \textbf{ChartQA} \\
\midrule
C\&P Consistency & 
\makecell{5.22 \\ (1.84$\times$10$^{-7}$)} & 
\makecell{4.75 \\ (2.18$\times$10$^{-6}$)} & 
\makecell{5.41 \\ (7.94$\times$10$^{-8}$)} & 
\makecell{2.36 \\ (7.22$\times$10$^{-3}$)} & 
\makecell{4.72 \\ (2.61$\times$10$^{-6}$)} \\
Cognitive Task & 
\makecell{4.69 \\ (2.76$\times$10$^{-6}$)} & 
\makecell{3.28 \\ (1.05$\times$10$^{-3}$)} & 
\makecell{3.32 \\ (9.17$\times$10$^{-4}$)} & 
\makecell{2.10 \\ (3.60$\times$10$^{-2}$)} & 
\makecell{4.55 \\ (5.70$\times$10$^{-6}$)} \\
Perceptual Task & 
\makecell{6.52 \\ (7.99$\times$10$^{-11}$)} & 
\makecell{4.49 \\ (7.43$\times$10$^{-6}$)} & 
\makecell{5.81 \\ (8.57$\times$10$^{-9}$)} & 
\makecell{2.53 \\ (1.28$\times$10$^{-2}$)} & 
\makecell{3.20 \\ (1.38$\times$10$^{-3}$)} \\
\bottomrule
\end{tabular}
\end{center}
\caption{Results of two-sided paired t-tests using InternVL2-2b, reported as t-statistics with p-values in parentheses.}
\label{tab:t-test}
\end{table*}

\clearpage

\begin{table*}
\begin{center}
\setlength\tabcolsep{4pt}
\begin{tabular}{lll}
\toprule
Prompt & \multicolumn{2}{p{13cm}}{You are tasked with determining whether the provided question-answer pairs   are examples of extractive question answering (Extractive QA).} \\
 &  &  \\
 & \multicolumn{2}{p{13cm}}{**You have been provided with the   following:**} \\
 & \multicolumn{2}{p{13cm}}{1. The document image.} \\
 & \multicolumn{2}{p{13cm}}{2. A list of question-answer pairs.} \\
 &  &  \\
 & \multicolumn{2}{p{13cm}}{**Here are the questions and   answers:**} \\
 & \multicolumn{2}{p{13cm}}{\{Question\_Answering\}} \\
 &  &  \\
 & \multicolumn{2}{p{13cm}}{**Definition of Extractive QA**} \\
 & \multicolumn{2}{p{13cm}}{In the domain of document   understanding, Extractive Question Answering (Extractive QA) refers to   systems that analyze and comprehend both the visual and textual information   within a document to directly extract answers to user queries from the   document's existing content. The answers are typically located in specific   sections of the document, eliminating the need for complex reasoning or the   generation of new content. Extractive QA emphasizes precise localization and   extraction of information to ensure the accuracy and verifiability of the   answers.} \\
 &  &  \\
 & \multicolumn{2}{p{13cm}}{**Non-Extractive QA Question   Types:**} \\
 & \multicolumn{2}{p{13cm}}{1. **Counting Questions:** These   require the system to count specific elements or occurrences within the   document, such as "How many times is the term 'machine learning'   mentioned in the report?"} \\
 & \multicolumn{2}{p{13cm}}{2. **Comparing Questions:** These   involve evaluating and contrasting different pieces of information within the   document, such as "Which department had a higher budget allocation in   Q2, Marketing or Sales?"} \\
 & \multicolumn{2}{p{13cm}}{3. **Causal Reasoning:** These   questions require understanding cause-effect relationships within the   document, such as "What caused the increase in operational costs?"} \\
 & \multicolumn{2}{p{13cm}}{4. **Synthesis Questions:** These   require summarizing or aggregating information from the document, such as   "Summarize the key findings of the annual report."} \\
 & \multicolumn{2}{p{13cm}}{5. **Inference Questions:** These   ask for conclusions based on implicit information within the document, such   as "What can be inferred about the company's market strategy from the   sales data?"} \\
 &  &  \\
 & \multicolumn{2}{p{13cm}}{**Your Task**} \\
 & \multicolumn{2}{p{13cm}}{For each question in the list,   determine whether it is an example of extractive QA based on the definition   provided.} \\
 &  &  \\
 & \multicolumn{2}{p{13cm}}{**Important:**} \\
 & \multicolumn{2}{p{13cm}}{- **Do not include any explanatory   content in your response.**} \\
 & \multicolumn{2}{p{13cm}}{- **Respond in the following format   for each question:**} \\
 & \multicolumn{2}{p{13cm}}{- If the question is extractive QA,   respond with: "Yes".} \\
 & \multicolumn{2}{p{13cm}}{- If the question is not extractive   QA, respond with: "No".} \\
 &  &  \\
 & \multicolumn{2}{p{13cm}}{**Example Response:**} \\
 & Q1: Yes &  \\
 & Q2: No &  \\
 & Q3: Yes &  \\
\midrule
Slots & Question\_Answering & List of question-answering annotations for the given   images. \\
\bottomrule
\end{tabular}
\end{center}
\vspace{-4mm}
\caption{Prompt for using GPT-4o to filter the questions that do not directly reference the text.}
\label{tab:prompt_4o_qa_filter}
\end{table*}
\clearpage

\begin{table*}
\begin{center}
\setlength\tabcolsep{4pt}
\begin{tabular}{lll}
\toprule
Prompt & \multicolumn{2}{p{13cm}}{You are tasked with identifying the locations of answers to multiple   questions about a document image.} \\
 &  &  \\
 & \multicolumn{2}{p{13cm}}{**You have been provided with the   following:**} \\
 & \multicolumn{2}{p{13cm}}{1. The document image.} \\
 & \multicolumn{2}{p{13cm}}{2. A list of questions along with   their corresponding answers.} \\
 & \multicolumn{2}{p{13cm}}{3. Text extracted from the document   image using an Optical Character Recognition (OCR) engine by a third party.} \\
 &  &  \\
 & \multicolumn{2}{p{13cm}}{**Here are the questions and   answers:**} \\
 & \multicolumn{2}{p{13cm}}{\{Question\_Answering\}} \\
 &  &  \\
 & \multicolumn{2}{p{13cm}}{**Here is the text extracted by the   OCR engine:**} \\
 & \{OCR\_Text\} &  \\
 &  &  \\
 & \multicolumn{2}{p{13cm}}{**Your task:**} \\
 & \multicolumn{2}{p{13cm}}{For each question in the list, first   determine whether the answer text can be found within the document image   based on the OCR-extracted text. If the answer is present, identify the box   ID(s) that contain the correct answer. Each answer appears **only once** in   the document image and may be entirely within a single box or span multiple   adjacent boxes, either horizontally or vertically. Include all relevant box   IDs that collectively constitute the answer. If the answer text cannot be   found in any box, indicate this as well.} \\
 &  &  \\
 & \multicolumn{2}{p{13cm}}{**It is important to emphasize that   you should identify only the boxes that contain the correct answer text, not   the boxes that are relevant to answering the question.** In other words, even   if a question explicitly mentions a specific box, if the answer text does not   appear in that box, it should not be considered.} \\
 &  &  \\
 & \multicolumn{2}{p{13cm}}{Keep in mind that you need to find   the box that semantically matches the answer, not just the box with the   answer text. This means you should fully consider all the information from   the document image, including images, text, layout, and style.} \\
 &  &  \\
 &  &  \\
 & \multicolumn{2}{p{13cm}}{**Important:**} \\
 & \multicolumn{2}{p{13cm}}{- **Do not include any explanatory   content in your response.**} \\
 & \multicolumn{2}{p{13cm}}{- **Respond in the following format   for each question:**} \\
 & \multicolumn{2}{p{13cm}}{- If you find the box(es) containing   the true answer, respond with: "Found {[}Box IDs{]}"} \\
 & \multicolumn{2}{p{13cm}}{- If you cannot find any boxes   containing the true answer, respond with: "Not Found"} \\
 &  &  \\
 & \multicolumn{2}{p{13cm}}{**Example Response:**} \\
 & \multicolumn{2}{p{13cm}}{Q1: Found {[}9, 12{]}} \\
 & \multicolumn{2}{p{13cm}}{Q2: Not Found} \\
 & \multicolumn{2}{p{13cm}}{Q3: Found {[}15{]}} \\
\midrule
Slots & Question\_Answering & List of question-answering annotations for the given images. \\
 & OCR\_Text & JSON-formatted OCR text for the given images. \\
\bottomrule
\end{tabular}
\end{center}
\caption{Prompt for using GPT-4o to locate $Box$ based on the OCR annotations of given image $x_I$.}
\label{tab:prompt_4o_locate}
\end{table*}
\clearpage

\begin{table*}
\begin{center}
\setlength\tabcolsep{4pt}
\begin{tabular}{lll}
\toprule
Prompt & \multicolumn{2}{p{13cm}}{You are given   several images with the page number indicated in the top left corner.} \\
\midrule
 & \multicolumn{2}{p{13cm}}{You will also receive a number of   independent question-answer pairs.} \\
 & \multicolumn{2}{p{13cm}}{For each question, your task is to   identify which numbered page provide the information needed to arrive at the   given answer.} \\
 & Note: &  \\
 & \multicolumn{2}{p{13cm}}{- Please identify which page these   key-value pairs are most likely to appear on.} \\
 & \multicolumn{2}{p{13cm}}{- Output only question-answer pair   id and its corresponding number. Format: Q1:number} \\
 & \multicolumn{2}{p{13cm}}{\{Question\_Answering\}} \\
 \midrule
Slots & Question\_Answering & List of question-answering annotations for the given images. \\
\bottomrule
\end{tabular}
\end{center}
\caption{Prompt for using GPT-4o to identify the correct page for information extraction on DeepForm.}
\label{tab:prompt_4o_selection}
\end{table*}

\begin{table*}
\begin{center}
\setlength\tabcolsep{4pt}
\begin{tabular}{lll}
\toprule
Prompt & \multicolumn{2}{p{13cm}}{You are asked to answer questions asked on a document image.} \\
 & \multicolumn{2}{p{13cm}}{The answers to questions are short   text spans taken verbatim from the document.} \\
 & \multicolumn{2}{p{13cm}}{This means that the answers comprise   a set of contiguous text tokens present in the document.} \\
 &  &  \\
 & \multicolumn{2}{p{13cm}}{Question: \{Question\}} \\
 &  &  \\
 & \multicolumn{2}{p{13cm}}{Directly extract the answer of the   question from the document with as few words as possible.} \\
 &  &  \\
 & Answer: &  \\
\midrule
Slots & Question & The question about the given image. \\
\bottomrule
\end{tabular}
\end{center}
\caption{Prompt for evaluating close-source MLLMs on cognitive task in DocVQA and DUDE.}
\label{tab:prompt_4o_docvqa_dude}
\end{table*}

\begin{table*}
\begin{center}
\setlength\tabcolsep{4pt}
\begin{tabular}{lll}
\toprule
Prompt & \multicolumn{2}{p{13cm}}{You are now working on DeepForm, a dataset for extracting text from visually   structured political ad receipts. This dataset focuses on five key fields:} \\
 &  &  \\
 & \multicolumn{2}{p{13cm}}{1. **contract\_num**: Contract number   (multiple documents can share the same number if a contract is revised)} \\
 & \multicolumn{2}{p{13cm}}{2. **advertiser**: Advertiser name   (often a political committee, but not always)} \\
 & \multicolumn{2}{p{13cm}}{3. **flight\_from / flight\_to**:   Start and end air dates for the ad (also known as "flight   dates")} \\
 & \multicolumn{2}{p{13cm}}{4. **gross\_amount**: Total amount   paid for the ads} \\
 &  &  \\
 & \multicolumn{2}{p{13cm}}{The answer always appears in the   document, but it may not match the exact words of the question or field name.   Provide a contiguous text span from the form, and include no additional   explanation besides the answer.} \\
 &  &  \\
 & \multicolumn{2}{p{13cm}}{Question: \{Question\}} \\
 &  &  \\
 & Answer: &  \\
 \midrule
Slots & Question & The question about the given image. \\
\bottomrule
\end{tabular}
\end{center}
\caption{Prompt for evaluating close-source MLLMs on cognitive task in DeepForm.}
\label{tab:prompt_4o_deepform}
\end{table*}

\begin{table*}
\begin{center}
\setlength\tabcolsep{4pt}
\begin{tabular}{lll}
\toprule
Prompt & \multicolumn{2}{p{13cm}}{You are now working on FUNSD, a dataset for form understanding in scanned   documents. These documents often contain text arranged in various sections,   tables, or multi-line blocks, and your goal is to extract the text that   directly answers each question. Your task is to return the contiguous text   snippet from the document that fully answers each question. The answer is   guaranteed to be present in the form image, so do not refuse. If the relevant   text spans multiple lines or rows in a table, ensure you include all of them   exactly as they appear. Avoid adding explanations or summarizing the text;   simply return a contiguous text snippet from the form that best addresses the   question.} \\
 &  &  \\
 & \multicolumn{2}{p{13cm}}{Question: \{Question\}} \\
 &  &  \\
 & Answer: &  \\
\midrule
Slots & Question & The question about the given image. \\
\bottomrule
\end{tabular}
\end{center}
\caption{Prompt for evaluating close-source MLLMs on cognitive task in FUNSD.}
\label{tab:prompt_4o_funsd}
\end{table*}

\begin{table*}
\begin{center}
\setlength\tabcolsep{4pt}
\begin{tabular}{lll}
\toprule
Prompt & \multicolumn{2}{p{13cm}}{You are analyzing a chart that may include numeric data, textual labels, and   visual features (e.g., bars, lines, colors). Below are some example questions   and answers from other charts—these examples are not from this chart. When   answering the current question, rely solely on the information in the chart   you are analyzing, and provide a concise answer based strictly on the chart’s   data. Avoid outside knowledge or extra explanations.} \\
 &  &  \\
 & \multicolumn{2}{p{13cm}}{Additionally, the question is   guaranteed to have an answer found in the chart. For numeric answers, remove   any commas or symbols (e.g., “\%”) unless specifically asked for. For   instance, “37,133” should be written as “37133” and “32.4\%” should be written   as “32.4.”} \\
 &  &  \\
 & \multicolumn{2}{p{13cm}}{Question: \{Question\}} \\
 &  &  \\
 & Answer: &  \\
 \midrule
Slots & Question & The question about the given image. \\
\bottomrule
\end{tabular}
\end{center}
\caption{Prompt for evaluating close-source MLLMs on cognitive task in ChartQA.}
\label{tab:prompt_4o_cahrtqa}
\end{table*}

\begin{table*}
\begin{center}
\setlength\tabcolsep{4pt}
\begin{tabular}{lll}
\toprule
Prompt & \multicolumn{2}{p{13cm}}{Analyze the provided image, which has a **single red box** containing text.  **Extract only**   the text inside this box, preserving the **original line order** from **top**   to **bottom**. If there are multiple lines, output them **separately**; if   there's just one line, output it **as is**. **Do not** include any text or   descriptions from outside the red box, and **do not** add any extra   punctuation, commentary, or code block markers. Return **only** the exact   text inside the red box.} \\
\bottomrule
\end{tabular}
\end{center}
\caption{Prompt for evaluating close-source MLLMs on perceptual task.}
\label{tab:prompt_4o_ocr}
\end{table*}

\begin{table*}
\begin{center}
\setlength\tabcolsep{4pt}
\begin{tabular}{lll}
\toprule
Prompt & \multicolumn{2}{p{13cm}}{**Task   Description**} \\
 &  &  \\
 & \multicolumn{2}{p{13cm}}{You are tasked with generating   potential OCR (Optical Character Recognition) error results based on the   provided list of question-answer (QA) pairs.} \\
 &  &  \\
 & \multicolumn{2}{p{13cm}}{**Provided Content:**} \\
 &  &  \\
 & \multicolumn{2}{p{13cm}}{**List of QA Pairs:**} \\
 & \multicolumn{2}{p{13cm}}{\{Question\_Answering\}} \\
 &  &  \\
 & \multicolumn{2}{p{13cm}}{**Your Task**} \\
 &  &  \\
 & \multicolumn{2}{p{13cm}}{For each QA pair, provide **3   possible OCR error results for the answer (A)**. **Each error result must   maintain a similar format, contain different content, must not be identical   to the original answer (A), and must be distinct from the other error results.**} \\
 &  &  \\
 & \multicolumn{2}{p{13cm}}{**Output Format**} \\
 &  &  \\
 & \multicolumn{2}{p{13cm}}{Please respond in **JSON** format   according to the structure provided below. Note that "error1,"   "error2," and "error3" are merely placeholders.} \\
\midrule
Slots & Question\_Answering & List of question-answering annotations for the given   images. \\
\bottomrule
\end{tabular}
\end{center}
\caption{Prompt for using GPT-4o to generate $y_C^-$ (Section \ref{sec:method}).}
\label{tab:prompt_4o_distrub}
\end{table*}



\end{document}